\documentclass[sn-basic]{sn-jnl}

\usepackage{graphicx}%
\usepackage{multirow}%
\usepackage{amsmath,amssymb,amsfonts}%
\usepackage{amsthm}%
\usepackage{mathrsfs}%
\usepackage[title]{appendix}%
\usepackage[table]{xcolor}%
\usepackage{textcomp}%
\usepackage{manyfoot}%
\usepackage{booktabs}%
\usepackage{algorithm}%
\usepackage{algorithmicx}%
\usepackage{algpseudocode}%
\usepackage{listings}%
\DeclareMathOperator*{\argmin}{argmin}

\usepackage{orcidlink}

\begin{document}
\rowcolors{2}{gray!25}{white}
\title[Article Title]{Active Learning in Genetic Programming: Guiding Efficient Data Collection for Symbolic Regression}


\author*[1]{\fnm{Nathan} \sur{Haut}\orcidlink{https://orcid.org/0000-0002-3989-6532}}\email{hautnath@msu.edu} 

\author[2]{\fnm{Wolfgang} \sur{Banzhaf}\orcidlink{https://orcid.org/0000-0002-6382-3245}}\email{banzhafw@msu.edu}

\author[1]{\fnm{Bill} \sur{Punch}\orcidlink{https://orcid.org/0000-0002-4493-6877}}\email{punch@msu.edu}

\affil*[1]{\orgdiv{CMSE}, \orgname{Michigan State University}, \orgaddress{\street{426 Auditorium Rd.}, \city{East Lansing}, \postcode{48824}, \state{Michigan}, \country{USA}}}

\affil[2]{\orgdiv{Computer Science}, \orgname{Michigan State University}, \orgaddress{\street{426 Auditorium Rd.}, \city{East Lansing}, \postcode{48824}, \state{Michigan}, \country{USA}}}



\abstract{ This paper examines various methods of computing uncertainty and diversity for active learning in genetic programming. We found that the model population in genetic programming can be exploited to select informative training data points by using a model ensemble combined with an uncertainty metric. We explored several uncertainty metrics and found that differential entropy performed the best. We also compared two data diversity metrics and found that correlation as a diversity metric performs better than minimum Euclidean distance, although there are some drawbacks that prevent correlation from being used on all problems. Finally, we combined uncertainty and diversity using a Pareto optimization approach to allow both to be considered in a balanced way to guide the selection of informative and unique data points for training.}




\maketitle

\section{Introduction}\label{sec1}

In applications of data science, the task of collecting and labelling data is often time-consuming and expensive. In some cases where data doesn't yet exist, it may be very expensive to run experiments to gather data, or possibly it could take long periods of time for experiments to complete. In these cases, it would be ideal to target specific experiments where maximal information will be gained, so fewer experiments have to be run to gain the desired insight into the system of study. In other cases, large masses of data may already exists, but the process of labelling the data is time-consuming. Here, it would be ideal to target a subset of samples, that when labelled, will provide the most information. To achieve these time-savings and cost reductions we can use machine learning (ML) not only to build models to describe these systems, but also to predict the information gained by each training sample. The process of using machine learning to iteratively select data to best inform machine learning model development is called {\it active learning}. 

\begin{figure}[h]
\centering
\includegraphics[width=12cm]{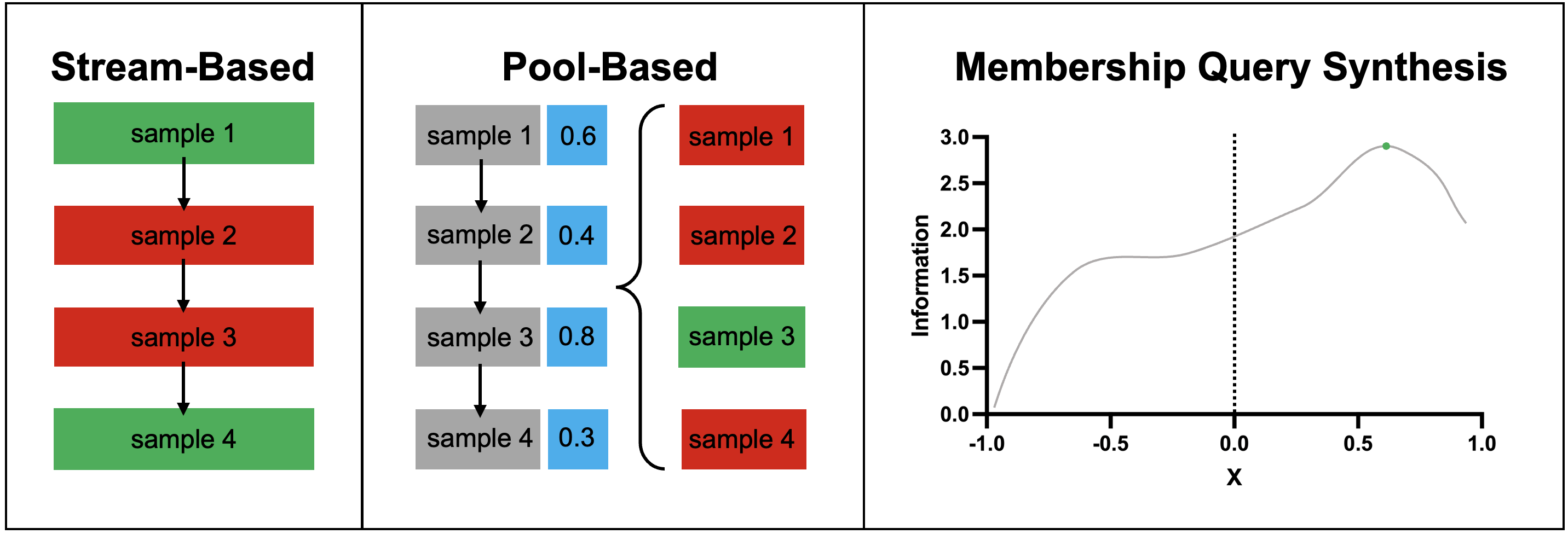}
\caption{The three main types of active learning: Stream-based, pool-based, and membership query synthesis are visually demonstrated. Stream-based approaches, shown on the left, search through the samples one at a time and either mark them for labelling or skip them. Green indicates a sample is found to be informative and is marked for labelling, red indicates a sample is skipped. Pool-based approaches, shown in the middle, assigns an information score to each potential training sample and the most informative sample is chosen to be labelled and added to the training set. Membership query synthesis, shown on the right, searches a space of potential points not yet collected while maximizing an information measure and selects a point to be synthesized and labelled that maximizes the information score. The selected point is indicated by the green circle, while the y-axis of the curve represents the informativeness measure and the x-axis is representative of the sample space.}
\label{fig:ALTypes}
\end{figure}

More specifically, active learning (AL) is a method used in conjunction with machine learning to actively select new training data with the goal of selecting data points that will maximally inform the machine learning model \citep{alintro}. Various forms of active learning exist, with three types  dominating: pool-based AL, stream-based AL, and membership query synthesis \citep{alsurvey}. Figure \ref{fig:ALTypes} shows a simple visual representation to compare the three methods of active learning. Pool-based and stream-based methods both have a set of training samples to choose from, with the goal of selecting and training on only a small subset of maximally informative cases. The key difference between pool-based and stream-based methods is that pool-based methods search over a set of data points for the ones that are most informative. Steam-based methods differ by checking each potential training case in order one-by-one and only admit them to the training set if data points are "informative". Membership query synthesis approaches do not have a set of already existing training samples to choose from, instead, they search a training space to find and synthesize new training data points that are expected to maximally inform the machine learning model. Once synthesized, a new data point is then labelled by the researcher via experimentation or expert knowledge. 

Active learning is a versatile method with uses ranging from effectively sub-sampling of data from a huge set for training, sampling of data with specific goals such as to maximize diversity, to guiding experimentation by suggesting experiments that will be most informative to the researcher in the model building process. It can be used to focus on interesting samples from large sets or to expand small data sets while minimizing data collection efforts.  For example, AL has recently been used to explore a space of 16 million potential catalysts to maximize the conversion rate of methane to methanol, which without active learning would not have been possible to search effectively within a reasonable time \citep{alChemUncertainty}. Active learning has also been shown to effectively sub-sample training data for identifying malware-infected PDF documents \citep{alCybersecurity}. The authors found that when using active learning they could reduce the training set size to 1/30-th of the original size, while maintaining the same 
performance as models trained on the whole set.

For a wide range of machine learning methods active learning approaches have been developed, e.g. for support vector machines or neural networks. In support vector machines, for instance, AL has been realized by computing the distance of all points to the separating hyperplane and selecting the point nearest the hyperplane to be labelled \citep{svmAL}. For neural networks, one AL variant has been to select points with minimum difference between the two most probable predicted labels \citep{al}. This distribution was defined as $M=P(l1|x)-P(l2|x)$, where $M$ is the margin between the two most probable labels, $l1$ is the most probably label for input $x$, and $l2$ is the second most probable label for input $x$.

In this contribution, we apply active learning strategies for genetic programming used in symbolic regression tasks. The goal is to exploit some of the features of GP, in particular its reliance on a population of models. More specifically, we want to utilize uncertainty and diversity measures in a model population context to accelerate the discovery of models (physics equations in our study). The idea is to look for disagreement among high-quality individuals in the population as a guide to locate informative data points to add to the training set. 

\section{Related Works}

Active learning methods for machine learning have shown to be very successful in applied settings to improve the method of labelling and collecting data with various machine learning types. AL has recently been demonstrated to significantly reduce the labelling efforts required for labelling data associated with identifying heart disease \citep{alHeartDisease}. The authors demonstrated that they could find more accurate models using fewer data points when compared to a random point selection strategy.

AL has been applied to genetic programming classification tasks as well. Using an ensemble of GP models, the models "vote" on the class of data pairs, and points are only labelled when the committee of developing models encounters pairs that can't be classified \citep{classify}. This was found to reduce the total effort needed to label training points, since only a subset had to be labelled before finding accurate models. Where GP training sets are large, AL has been successfully applied by selecting sub-samples to be used for training \citep{gpFitnessAL,sample}. In \citep{sample} AL is performed by segmenting the data into smaller blocks and training the models using one randomly selected block at a time using uniform probability. As training continues, bias is introduced into the probability by increasing the tendency to select blocks that haven't been seen in a while, as well as blocks where the models performed poorly during training. AL for sub-sampling with genetic programming was found to decrease training times to find better binary classification models by an order of magnitude \citep{sample}. In \citep{gpFitnessAL} subsets were selected by dynamically developing a fitness case topology that could be used to create minimally related subsets of data. In this context, the strength of a relationship between two training cases was indicated by the number of individuals that were able to solve both training cases. 

In the discovery of biological networks AL methods have also been employed successfully \citep{bio}. Several different approaches were explored by the authors for determining which new data points would be maximally informative for a wide range of machine learning models, including Boolean networks, causal Bayesian networks, differential equation models, etc. One approach the authors explored was the maximum difference method in which two best-fit models are chosen and a new data point is selected where those two best-fit models have the largest difference in predictions. They also examined entropy score maximization. In that method a new data point is selected that maximizes an entropy score, where entropy can be thought of as the amount of information to be gained by gathering that data point. The entropy score $H_e$ is computed as follows:
\begin{equation*}
   H_e=-\sum _{x=1}^{x_e} \frac{e_x}{|M|}\log _2\frac{e_x}{|M|}
\end{equation*}
where $M$ is the set of Boolean networks, $x_e$ is the number of network states for a given data point, and $e$ is the set of all potential data points.  

In chemical engineering AL has been applied to expedite a reaction screening process by only selecting a subset of maximally informative experiments to complete rather than by exhaustively performing all possible experiments \citep{chem}. This was done by training neural networks and using them to select a subset of experiments that maximized the information gain. Maximal information gain was determined by looking at the standard deviation of an ensemble of neural networks.  


Kotanchek et al. (\citeyear{trustable}) used genetic programming for active design of experiments, where models developed by a GP system are used to find optimal conditions in a system of study. Active design of experiments is an application of active learning, where it has the goal of designing experiments that have specific properties or yield maximal information. The authors proposed to employ ensembles of models from symbolic regression to find regions of uncertainty in order to gather new data with high information content. 
While this method has been proposed for how an active learning method using model ensembles could be applied to GP for symbolic regression, there has yet to be any research showing how active learning methods affect the performance of GP symbolic regression tasks or how the method to quantify uncertainty affects the quality of points selected for inclusion in the training data. As well, it is yet to be shown that this idea of selecting an ensemble from a model population and searching for points of high uncertainty or disagreement among models is generalizable to any machine learning method where a population of models is available.

\section{Methods}

We compare two classes of active learning: uncertainty and diversity based. The implementations are described in detail below. We use two random sampling methods as a baseline to compare the performance of the active learning methods. The key features of the GP system we used, StackGP, are also discussed. 

\begin{figure}[h]
\centering
\includegraphics[width=12cm]{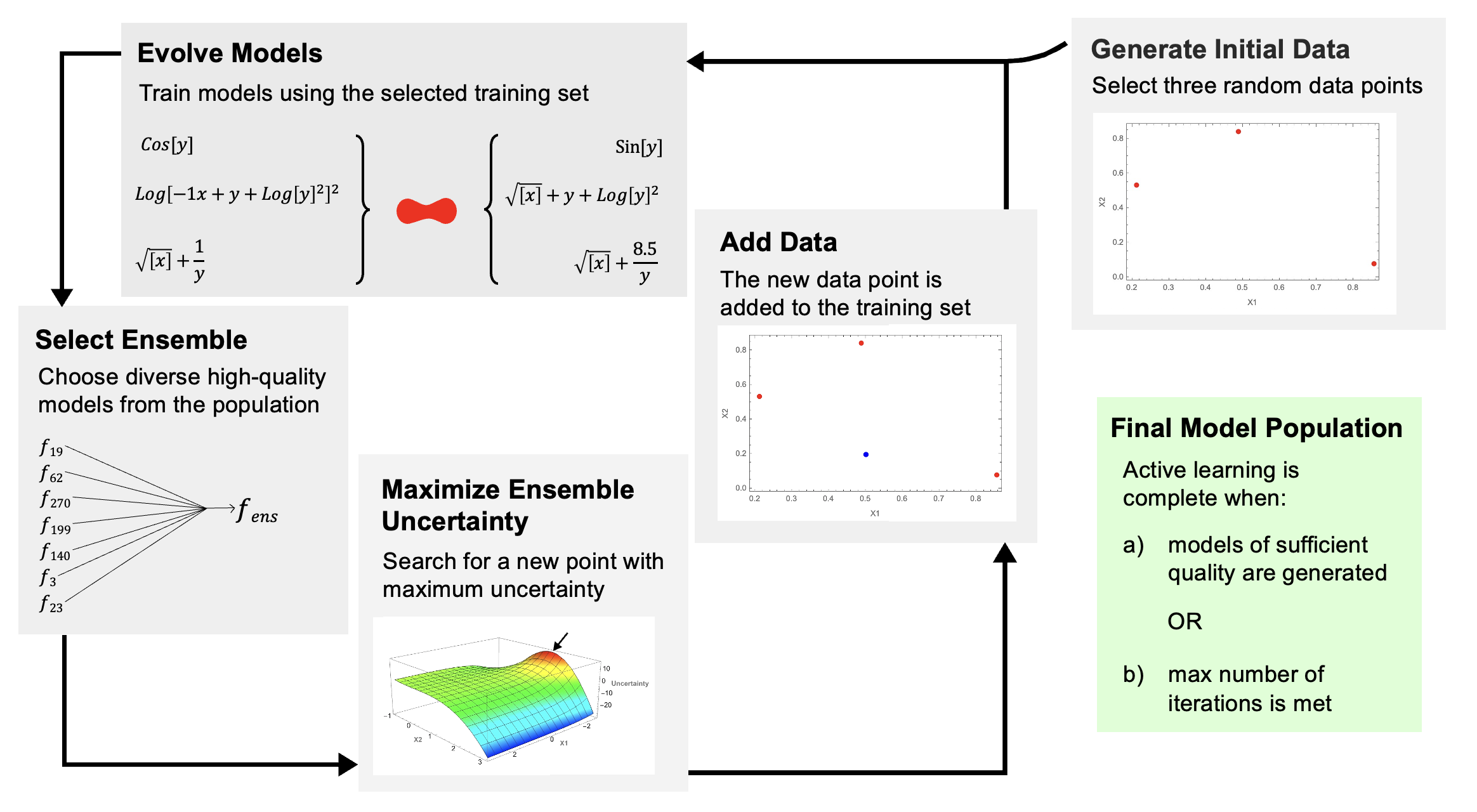}
\caption{An overview of the iterative active learning approach. It begins with an initially randomly selected dataset. It then iteratively evolves models and selects new training points that maximize uncertainty of an ensemble of models. By maximizing ensemble uncertainty to select new training samples, points with relatively high information content are added to the training set each iteration. }
\label{fig:ALOverview}
\end{figure}

\subsection{Active Learning}
Two general types of active learning were implemented to work with StackGP for the purpose of accelerating the development of models to fit physics data from the Feynman Symbolic Regression Dataset (\citeauthor{data}). The first type of active learning explored was uncertainty-based, a model-driven approach to active learning, where an ensemble of diverse, high-quality models from a population was used to search for regions in the search space where there was high uncertainty or disagreement between the models. The second type of active learning explored was diversity-based active learning, where new points are selected that differ maximally from the points already in the training sample. This second type of active learning is a data-driven approach rather than a model-driven approach. The first type of active learning is summarized in Figure \ref{fig:ALOverview}. 

Both types of active learning methods were implemented to determine how they each impact the success of evolution in genetic programming symbolic regression tasks. Several different uncertainty and diversity metrics are implemented to determine their respective impact on the success of the task. Success of active learning by maximizing uncertainty would indicate that the diversity of the population can be utilized to guide the collection of informative data. Success of diversity sampling would indicate that GP symbolic regression model development benefits from improved data sampling.

\subsubsection{Maximizing Uncertainty}

Several different uncertainty metrics were explored to determine how different measures impact the success of active learning, and which approach would generally work best. As an overview, each approach begins by selecting an ensemble of models using the same method, then a function  that uses the specific uncertainty metric along with the ensemble and current training set is created. This function is then fed to an optimizer to search for regions of relatively high uncertainty. The most uncertain point found is then returned and selected to be added to the training set. In total, there were 6 different uncertainty maximization approaches tested which varied in how they quantified disagreement, whether outlier predictions were considered, and which optimizer was used. The steps and methods will be described in greater detail below and the entire process is depicted in Algorithm \ref{alg:ALU}. 

Generating the ensemble is the first step in uncertainty-based active learning. The goals for generating the ensemble were to capture diverse, high-quality individuals from the population while keeping the size of the ensemble relatively small so that the computational cost of optimizing uncertainty is reasonable. The diversity goal is essential to the success of active learning since disagreement between models is a necessary requirement. The method chosen to capture both diversity and quality from the model population works by clustering the training data using the input space and selecting a model that best fits each cluster, ensuring no model is selected more than once. If a model is already selected by another cluster, the next best unselected model is chosen. The minimum number of clusters is set to 3 and the maximum is set to 10. Thus, 3-10 models are chosen for inclusion in an ensemble. Data clustering was chosen with the intent to capture diversity by focusing on models that have biases for different regions of the training space. Quality in the population would be captured since only models with the best fitness were selected for each cluster. The algorithm to generate the ensemble is described in detail in Algorithm \ref{alg:ensembleAlg}. 

\begin{algorithm}
\scriptsize{\tiny}
\caption{Ensemble generation process to select diverse high-quality models. }\label{alg:ensembleAlg}
\begin{algorithmic}
\Procedure{EnsembleSelect}{$models$,$trainingData$,$responseData$}
  \State $selectedModels \gets []$ \Comment{Initialize ensemble}
  \State $nClusters \gets min(len(trainingData),10)$ \Comment{Determine number of clusters}
  \State $clusters \gets KMeans(nClusters).fit\_predict(trainingData)$
  \For{$i=0$;$i++$; $i<nClusters$} \Comment{Loop over data clusters}
    \State $modelErrors \gets computeError(models,clusters[i])$
    \State $sortedModels \gets sortBy(models,modelErrors)$
    \State $j=0$
    \While{$sortedModels[j]$ in $selectedModels$} \Comment{Find best unselected model}
        \State $j++$
    \EndWhile
    \State $selectedModels=join(selectedModels,sortedModels[j]$ \Comment{Add to ensemble}
  \EndFor
  \State return $selectedModels$ \Comment{Return ensemble}
\EndProcedure

\end{algorithmic}
\end{algorithm}

The second step of this method is to utilize the specified uncertainty function with both the current training data and the selected ensemble. The function is then given to the optimizer with the search space boundaries to find a point of relatively high uncertainty. In the case that an already selected point is re-selected, a new search is initiated within a random sub-region until a unique point is added. This ensures that new information is added in each iteration to the training set. 

The two methods used for optimization were Scipy Optimize's minimize and differential evolution (\citeauthor{scipy.optimize.minimize,scipy.optimize.diff}). 

In total 5 different uncertainty metrics were used, shown by Equations \ref{eq:StdMean} to~\ref{eq:De}, where Equation \ref{eq:De} is used twice, once with Scipy's minimize function for optimization, and a second time with Scipy's differential evolution function for optimization. 

\begin{equation}
\label{eq:StdMean}
    \Delta = \frac{\text{Std}(\text{EnsembleResponses})}{\text{Mean(Abs(}\text{EnsembleResponses}))}
\end{equation}

\begin{equation}
\label{eq:TrStdTrMean}
    \Delta = \frac{\text{TrimmedStd}(\text{EnsembleResponses},0.3)}{\text{TrimmedMean(Abs(}\text{EnsembleResponses}),0.3)}
\end{equation}

\begin{equation}
\label{eq:StdTrMean}
    \Delta = \frac{\text{Std}(\text{EnsembleResponses})}{\text{TrimmedMean(Abs(}\text{EnsembleResponses}),0.3)}
\end{equation}

\begin{equation}
\label{eq:Std}
    \Delta = \text{Std}(\text{EnsembleResponses})
\end{equation}

\begin{equation}
\label{eq:De}
    \Delta = \text{DifferentialEntropy}(\text{EnsembleResponses})
\end{equation}



\begin{algorithm}
\scriptsize{\tiny}
\caption{Active Learning Process Using Uncertainty}\label{alg:ALU}
\begin{algorithmic}
\State $Training Data \gets 3StartingPoints$                \Comment{Generate initial random training data}
\State $Models \gets RandomModels$
\Comment{Generate initial random models}
\State $Models \gets Evolve(TrainingData, Models)$                  \Comment{Train models on starting data}
\While{$Best Model Error \neq 0$}                           \Comment{While perfect model not found}
    \State $Ensemble \gets EnsembleSelect(Models)$.         \Comment{Select ensemble of models}
    \State $New Point \gets MaxUncertainty(Ensemble)$  \Comment{Find point of max uncertainty}
    \If{$NewPoint \subset TrainingData$}                  \Comment{If point already selected}
        \State $NewPoint \gets MaxUncertainty(SubSpace(Ensemble))$\Comment{Search a subspace}
    \EndIf
    \State $Training Data \gets Append(Training Data, New Point)$ \Comment{Add new point}
    \State $Models \gets Evolve(Training Data, Models)$           \Comment{Evolve new models with new data using best models to seed evolution}
\EndWhile
\end{algorithmic}
\end{algorithm}

\subsubsection{Point Diversity}
A data-driven active learning approach was also explored, aiming to maximize data diversity rather than maximize ensemble uncertainty. The goal was to determine if GP evolution for symbolic regression tasks would benefit significantly from improved sampling of the data for training. Two different metrics were used to quantify diversity: point distance and point correlation. Point distance was implemented by measuring both the minimum and average Euclidean distance to all points in the training set. Point correlation was defined as the average correlation to all points in the training set. When selecting a new point, the goal was to either maximize the distance or minimize the correlation to the current training set.  

To minimize the correlation when selecting a new point, Pearson's $R^2$ was computed between each point and the potential new point. The equation for computing Pearson's $R$ is shown in Equation \ref{eq:corr1}. Here $y$ represents the new training point, $\hat{y}$ represents a point already in the set, and each instance $i$ represents the value in the ith dimension of the point. The overall method for computing the joint correlation of a new point to the training set is summarized in Algorithm \ref{alg:JointCorr}. 

\begin{equation}
\label{eq:corr1}
    R = \frac{\sum_{i=1}^N (y_i - \bar{y}) (\hat{y}_i - \bar{\hat{y}})}{\sqrt{\sum_{i=1}^N (y_i - \bar{y})^2 \times \sum_{i=1}^N (\hat{y}_i - \bar{\hat{y}})^2}}
\end{equation}

\begin{algorithm}
\scriptsize{\tiny}
\caption{}
\label{alg:JointCorr}
\begin{algorithmic}[1]
\Procedure{JointCorrelation}{$trainingSet$,$newPoint$}
  \State $r2Values \gets [PearsonR(trainPt,newPoint)^2$ for $trainPt$ in $trainingSet$] \Comment{$R^2$ vals}
  \State $avgCorr \gets mean(r2Values)$ \Comment{Compute average correlation}
  \State Return $avgCorr$
\EndProcedure

\end{algorithmic}
\end{algorithm}

\begin{algorithm}
\scriptsize{\tiny}
\caption{Active Learning Process Using Diversity}\label{alg:ALD}
\begin{algorithmic}
\State $Training Data \gets 3StartingPoints$                \Comment{Generate initial random training data}
\State $Models \gets RandomModels$
\Comment{Generate initial random models}
\State $Models \gets Evolve(TrainingData, Models)$                  \Comment{Train models on starting data}
\While{$Best Model Error \neq 0$}                           \Comment{While perfect model not found}
    \State $New Point \gets MaxDiversity(Training Data)$  \Comment{Find point of max uncertainty}
    \If{$NewPoint \subset TrainingData$}                  \Comment{If point already selected}
        \State $NewPoint \gets MaxUncertainty(SubSpace(TrainingData))$\Comment{Search a subspace}
    \EndIf
    \State $Training Data \gets Append(Training Data, New Point)$ \Comment{Add new point}
    \State $Models \gets Evolve(Training Data, Models)$           \Comment{Evolve new models with new data using best models to seed evolution}
\EndWhile
\end{algorithmic}
\end{algorithm}

\subsubsection{Benchmark Testing}

Each active learning approach was compared on a benchmark set of 35 of the 100 equations from the Feynman Symbolic Regression Dataset \citep{feynman}. These particular 35 problems were selected since they were thought to be most appropriate for a study in active learning. In a previous study, 37 other of the 100 equations were consistently found to need just 3 data points to be solved when using StackGP \citep{ALStackGPGECCO}. This would render active learning useless in such cases. 
The remaining 28 equations generally required all the data points up to 1000 (as we tested) to reach moderate results, so it did not seem that this type of active learning, adding one point at a time, would be appropriate for those problems

\subsection{StackGP}

StackGP is a stack-based genetic programming implementation in Python \citep{ALStackGPGECCO} and is available here (\citeauthor{StackGPRepo}).

\subsubsection{Model Structure}

Similar to PushGP (\citeauthor{pushgp}), StackGP models use multiple stacks, where the model evaluation is driven by an operator stack while variables, constants, and other data types are stored on separate stacks. For symbolic regression tasks, we have a total of 2 stacks, the operator stack and the variables/constants stack. 

\subsubsection{Correlation Fitness Function}

Unlike many symbolic regression implementations that use (R)MSE as the fitness function, we employ correlation as the fitness function, together with a linear scaling post-processing step. This was shown to perform better than (R)MSE in earlier work \citep{corrFitGPTPip}. The fitness is optimized during search by first maximizing $R^2$, which is computed using Equation \ref{eq:corr}, where $N$ is the number of data points $i$, $y_i$ is the target output, and $\hat{y}_i$ the output calculated by the model. 
\begin{equation}
\label{eq:corr}
    R = \frac{\sum_{i=1}^N (y_i - \bar{y}) (\hat{y}_i - \bar{\hat{y}})}{\sqrt{\sum_{i=1}^N (y_i - \bar{y})^2 \times \sum_{i=1}^N (\hat{y}_i - \bar{\hat{y}})^2}}
\end{equation}

The search is then completed using a post-processing step, which aligns the resulting models via a simple linear regression step (eq. \ref{eq:linscale}), minimizing 
\begin{equation}
\label{eq:linscale}
    \argmin_{a_0, a_1} \sum_{i=1}^N (|y_i - (a_1 \hat{y_i} + a_0)|)
\end{equation}

\subsubsection{Algorithm}

An overview of the algorithm is shown in Algorithm \ref{alg:StackGP}. The parameters used to run the algorithm are shown in Table \ref{tab:parameters}. Note that crossover and mutation calls in the algorithm are simplified and actually represent applying crossover and mutation to the correct fractions of models as shown in the parameters. 

Crossover is performed using a 2-point crossover operator where two points are selected in the operator stack of each parent and the operators, along with the associated variables and constants between the points, are swapped between the parents. Mutation has several different forms, each occurring with equal probability: random replacement of a variable, random replacement of an operator, pushing a random operator to the top of the operator stack and pushing variables/constants to the second stack when arity is greater than 1, popping a random number of operators off the operator stack and the correct number of variables/constants off the second stack, inserting a single operator at a random position in the stack, 2-point crossover with a random model, and appending a random operator to the bottom of the operator stack. There is then a repair mechanism that will push variables and constants to the top of the second stack if - after mutation - there are not enough items in the variable/constant stack for the operators. 

The tournament selection method used was Pareto tournament selection, where correlation and complexity were the two objectives. Complexity was measured as the combined stack lengths. 

\begin{table}

\label{tab:parameters}
\begin{tabular}{ll} 
\hline
\rowcolor{gray!50}
Parameter & Setting\\
\hline
 Mutation Rate & 79 \\ 
 Crossover Rate & 11 \\
 
 Spawn Rate & 10\\
 
 Elitism Rate & 10\\
 
 Crossover Method \hspace{5mm}  & 2 Pt.\\
 
 Tournament Size & 5\\
 
 Population Size & 300\\
 
 Selection Rate & 20\\

 Parallel Runs & 4 \\

 Generations & 1000 \\
\hline
\end{tabular}
\caption{StackGP \& Active learning Parameter Settings}
\end{table}

\begin{algorithm}
\caption{StackGP Search Algorithm}
\label{alg:StackGP}
\begin{algorithmic}[1]
\Procedure{Evolve}{$trainingData$,$models$}
  \For {generations 1 to 100}
  \State $models \gets setModelQuality(models,trainingData)$
  \State $newPop \gets ElitismSelection(models,20\%)$
  \State $models \gets tournamentSelection(models)$
  \State $newPop \gets newPop + crossover(models)+mutation(models)$
  \State $newPop \gets newPop + randomNewModels$
  \State $newPop \gets deleteDuplicates(newPop)$
  \State $models \gets newPop$
  \EndFor
  \State $alignedModels \gets alignment(models,trainingData)$
  \State Return $alignedModels$
\EndProcedure

\end{algorithmic}
\end{algorithm}

\subsection{Random Sampling}

As a baseline, we used random sampling of data points from uniform and normal distributions to determine if an active learning method improves learning progress over a naive sampling of training data. Uniform random sampling was chosen since it is a commonly used distribution and would likely be a first choice for naively sampling data. A normal distribution was selected since according to the central limit theorem, normal distributions tend to arise in nature, so a data set sampled from natural processes would likely be a normal distribution. 

To create a fair comparison against the active learning methods, a simple substitution was made where instead of using active learning to maximize uncertainty or diversity, a random point was added in each iteration. Beyond that substitution, the algorithm remains the same. 

The normal distribution for each variable was defined using the midpoint between the sampling bounds as the mean and $1/6$ of the difference between the upper and lower bounds as the standard deviation. This places 99.8\% of the distribution between the upper and lower bounds of each variable. If a point is sampled beyond a boundary it is adjusted to be on the boundary instead, although this is unlikely to occur frequently.

\section{Results and Discussion}

Several different approaches for computing uncertainty and diversity were compared using the Feynman Symbolic Regression Dataset. We then combine diversity and uncertainty using a Pareto optimization approach and compare that multi-objective method to using both uncertainty and diversity alone. The Pareto approach is then tested on two additional benchmark problems from the SRBench benchmark set.

\subsection{Active Learning Uncertainty Sampling}

The results of comparing the different uncertainty-based active learning methods are shown in Figures \ref{fig:uncertainty} and \ref{fig:uncertaintyCount} and the full table is in the Appendix as Table \ref{tab:uncTab}. 
Figure \ref{fig:uncertainty} uses uniform random sampling as the baseline for comparison, shown as the blue line in the figure. We also include normally distributed random sampling for comparison as the red distribution. The results show that the relative uncertainty measures, where we divide by the mean or trimmed mean, do not consistently perform better than uniform random sampling. The non-relative uncertainty measure performed well more consistently with the methods that use differential entropy performing best. The fact that standard deviation alone as an uncertainty metric performs consistently well is appealing since it is very cheap and easy to implement relative to some of the others. Differential entropy when using differential evolution as the optimizer performed best. The fact that differential evolution as the optimizer worked best with differential entropy likely indicates that the surface is highly non-convex, so differential evolution was better able to search the uncertainty space. 

\begin{figure}[h]
\centering
\includegraphics[width=12cm]{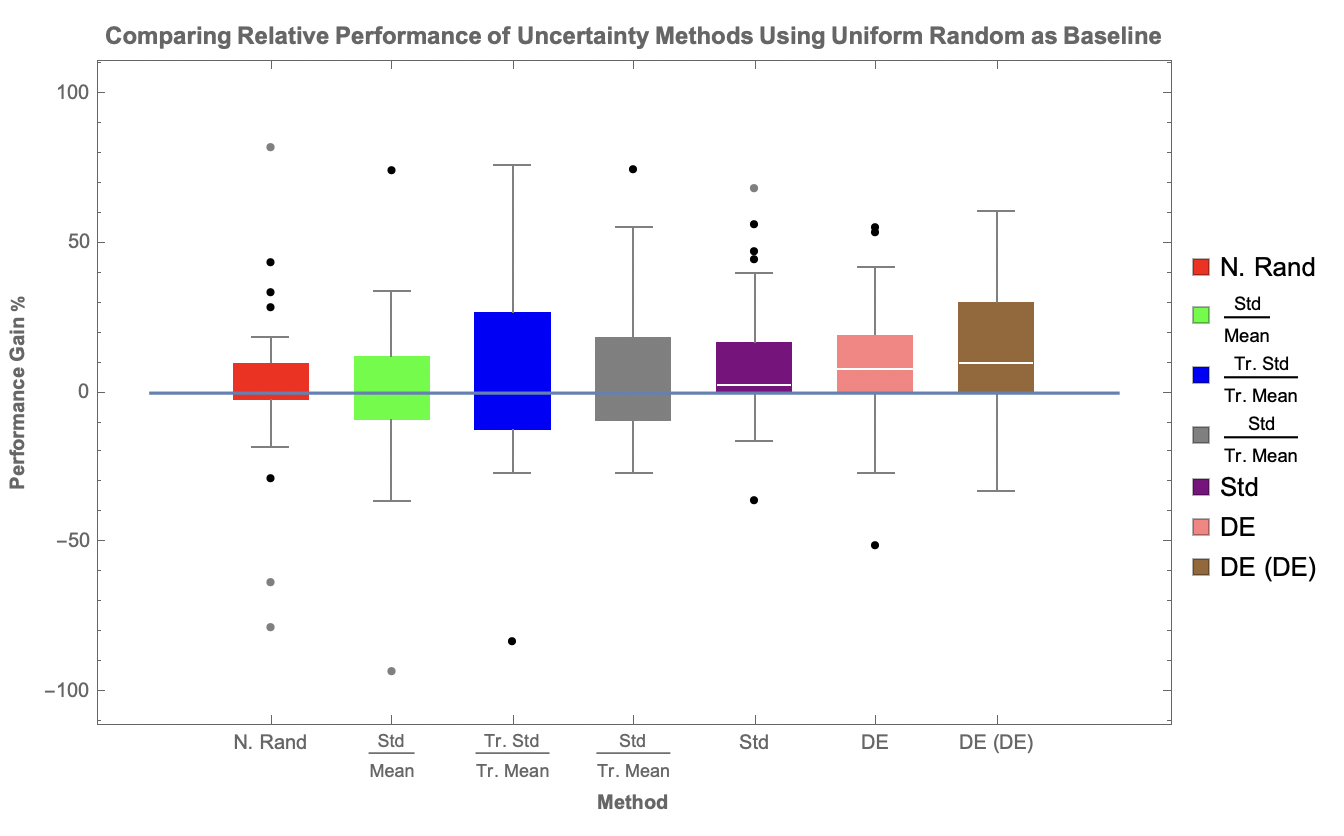}
\caption{{\bf Comparing Relative Performance of Uncertainty Methods Using Uniform Random Selection as Baseline.} Shown here are the performance differences of AL uncertainty methods compared to uniform random selection as the baseline (blue line) and normally distributed random selection (red distribution). We see that using the relative uncertainty measures where we divided by the mean we get inconsistent performance, sometimes performing much better than random but sometimes performing much worse. The non-relative approaches all consistently perform better than random selection with the methods that use differential entropy performing best. Using differential entropy with differential evolution (brown) we observe the best performance. The distributions represent the median performances of 100 independent runs across all test problems. For completeness, there is one point not shown for the std/tr. mean approach that is around -200.}
\label{fig:uncertainty}
\end{figure}

\begin{figure}[h]
\centering
\includegraphics[width=12cm]{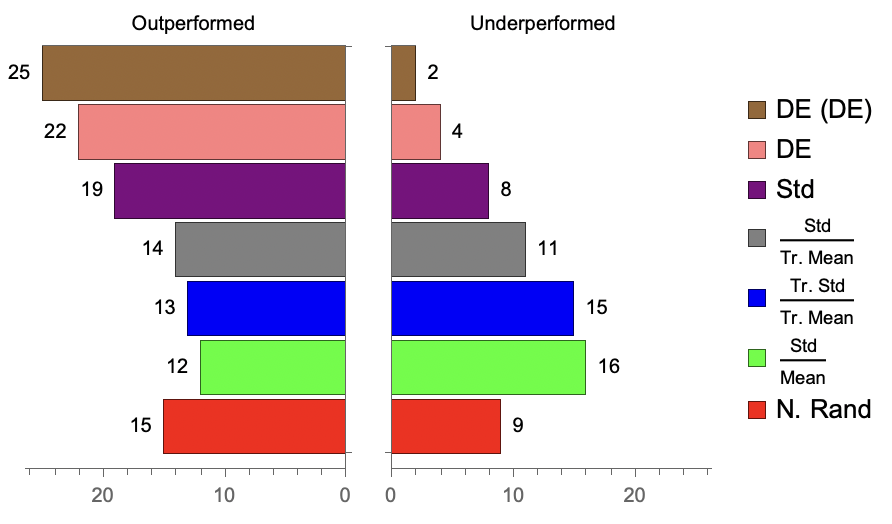}
\caption{{\bf Comparing Performance of Uncertainty Methods Against Uniform Random Selection.} Each method is compared to uniform random sampling and the number of times that the method outperforms and underperforms is reported. The number of times each method outperforms is shown on the left and the number of times each method underperforms is shown on the right. Outperforming means that a method used fewer points than uniform random sampling. Underperforming means that it required more points. Ties are not counted but can be easily determined by taking the difference of 35 and the two values reported. The results show that the methods that use differential entropy work well most consistently, outperforming more frequently and underperforming infrequently. We can also see that the relative uncertainty measures were very inconsistent in their performance.  }
\label{fig:uncertaintyCount}

\end{figure}

Figure \ref{fig:uncertaintyCount} compares the performance of each method against uniform random sampling for each problem and displays the number of times each method outperforms or underperforms random sampling. If a method outperforms random sampling that means that the method required fewer points to solve a problem. If a method underperforms random sampling that means that the method required more points to solve a problem. The results show that the methods using differential entropy work best, outperforming in the most number of cases and underperforming in the fewest number of cases. The differential entropy method that used differential evolution as the optimizer worked better than just using differential entropy with SciPy Optimize's minimize function. This indicates that differential evolution was able to search the uncertainty surface more effectively. The results also show that the relative uncertainty methods that divided the mean or trimmed mean were not consistent in their performance, frequently having a similar number of cases where the methods outperformed and underperformed. 

We see that the relative measures sometimes perform well and sometimes perform poorly, but on average they are centered around the baseline performance. The original assumption was that the relative uncertainty measures would be appealing since it was thought that they would reduce a bias towards selecting points where the predicted response is larger and thus naturally leads to wider distributions of the ensemble. This may have been the case occasionally where those methods did perform much better than uniform random sampling, but they ware not consistent. Looking at their formulations there is a risk of selecting points where the mean is near 0 which results in asymptotic behavior of the uncertainty function. 

Considering the results, we also see that of the two random sampling methods, normally distributed random sampling seems to perform a bit better than uniform sampling. This indicates that if a researcher does not want to use active learning to guide their data collection, they would typically be better off using a normal distribution than a uniform distribution for their samples.

\subsection{Active Learning Diversity Sampling}

\begin{figure}[h]
\centering
\includegraphics[width=12cm]{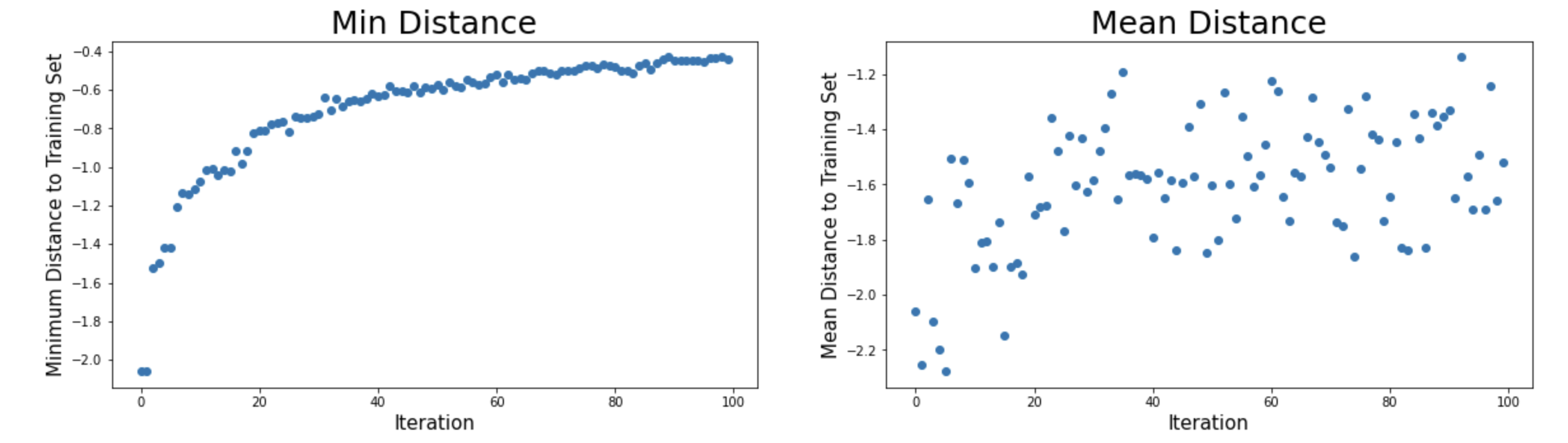}
\caption{Comparing minimum Euclidean distance against mean Euclidean distance as a diversity metric. Here minimum distance is used to select the next point in the set and both metrics of those points are displayed. We can see that there is little correlation between the two metrics indicating they provide different information. The $R^2$ between these two metrics on these points is just ~0.37. The Spearman Rho, rank-correlation, is also low at ~0.44.}
\label{fig:minmean}
\end{figure}

The different metrics for determining point diversity were compared to determine if there are clear differences in what they are measuring and also to ensure there aren't any obvious flaws with any of the metrics. When comparing minimum distance and average distance an initial randomly generated training set with 3 data points in 3 dimensions was generated. Figure \ref{fig:minmean} shows the comparison where new points were selected iteratively to add to the training set using the minimum distance metric for selection. We can see that the correlation, $R^2$ is actually pretty weak between the two, indicating they are providing different measures. As well, we recorded the Spearman Rho, rank-correlation, since that indicates if the methods are ranking points similarly or not. If methods rank points similarly, then they would likely not provide unique information if used as a diversity metric. It was found that the Spearman Rho was 0.44, which means that the two methods are ranking points differently and could provide unique information. 

\begin{figure}[h]
\centering
\includegraphics[width=12cm]{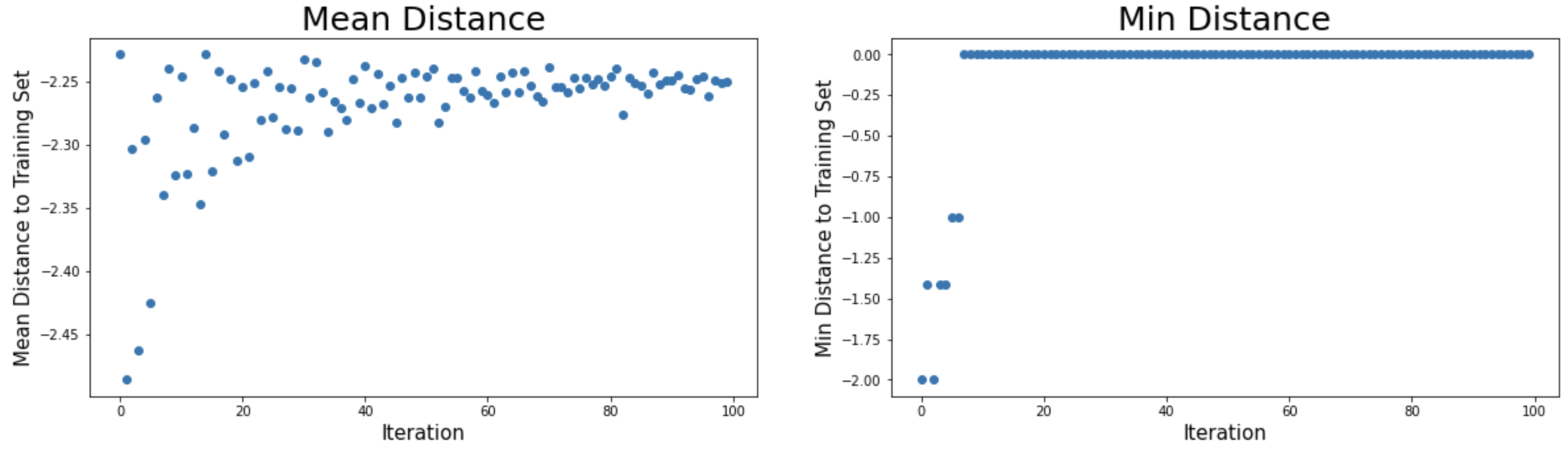}
\caption{Comparing minimum Euclidean distance against mean Euclidean distance as a diversity metric. Here mean distance is used to select the next point in the set and both metrics of those points are displayed. We can see that when mean distance is used to select new points, we get many points with a minimum distance of 0. This indicates that we are very frequently reselecting points already in the set. This shows that minimum distance is a better metric than mean distance. }
\label{fig:meanmin}
\end{figure}

\begin{figure}[h]
\centering
\includegraphics[width=12cm]{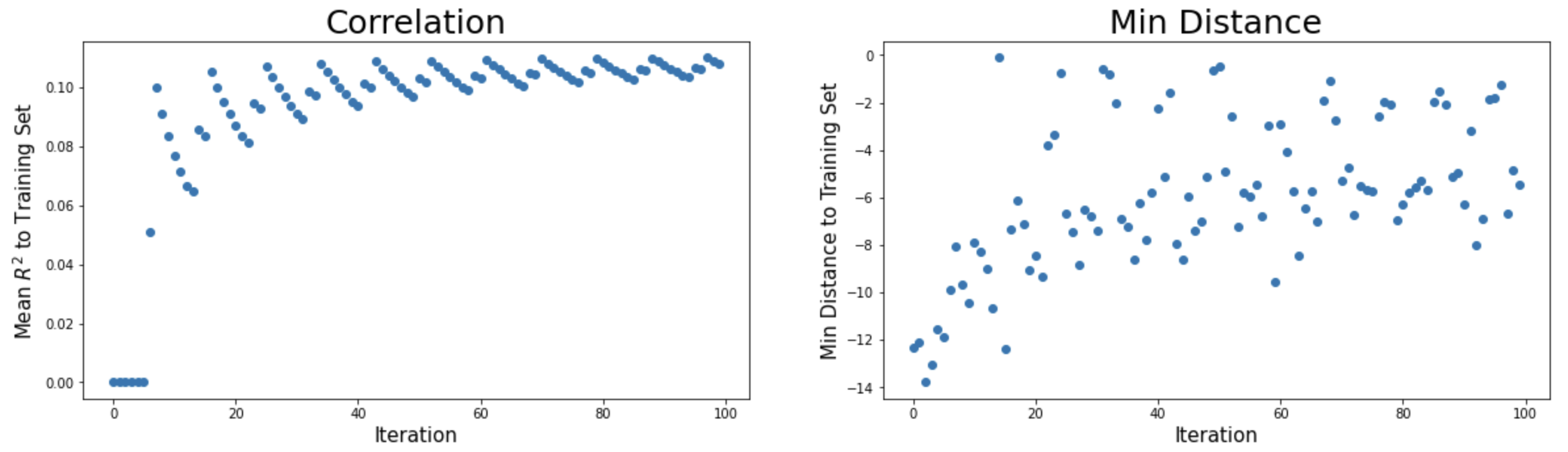}
\caption{Comparing minimum Euclidean distance against mean point correlation as a diversity metric. Here minimizing mean correlation is used to select the next point in the set and both metrics of those points are displayed. We can see that there appears to be a weak positive correlation between the two, indicating that they provide some of the same information but are not the same, so may have different advantages. It is also promising that the minimum distance shows that we are not reselecting points already in the training set. Comparing the metrics for these points we get an $R^2$ of ~0.35 and a Spearman Rho of ~0.33.}
\label{fig:corrmin}
\end{figure}

To further compare the minimum and mean distance metrics, the analysis was flipped, such that mean distance was used to select new points and both metrics were recorded on the selected points. These results are shown in Figure \ref{fig:meanmin}. Here it becomes obvious that mean distance is not a good metric since the minimum distance metric indicates that we are repeatedly selecting points already in the set. This is shown by the consistent minimum distance value of 0 after around 10 iterations. This result led to mean distance being thrown out as a potential choice of metric. 

Minimum distance and correlation were also compared to determine if they provide unique measure of diversity. The results are shown in Figure \ref{fig:corrmin}. For this analysis, lack of correlation to the training set was used to select new points and both metrics were recorded. This analysis was slightly different than the previous ones since for this problem the points were embedded in a 10 dimensional space instead of just 3. The results show that the two metric do provide unique information since an $R^2$ value of 0.35 and a Spearman Rho value of 0.33 were recorded, which are both low. Since these metrics were determined to provide unique information without any clear flaws both were included to be explored, with the one limitation that correlation as a diversity metric could not be used on problems of less than 3 dimensions. 

\begin{figure}[h]
\centering
\includegraphics[width=12cm]{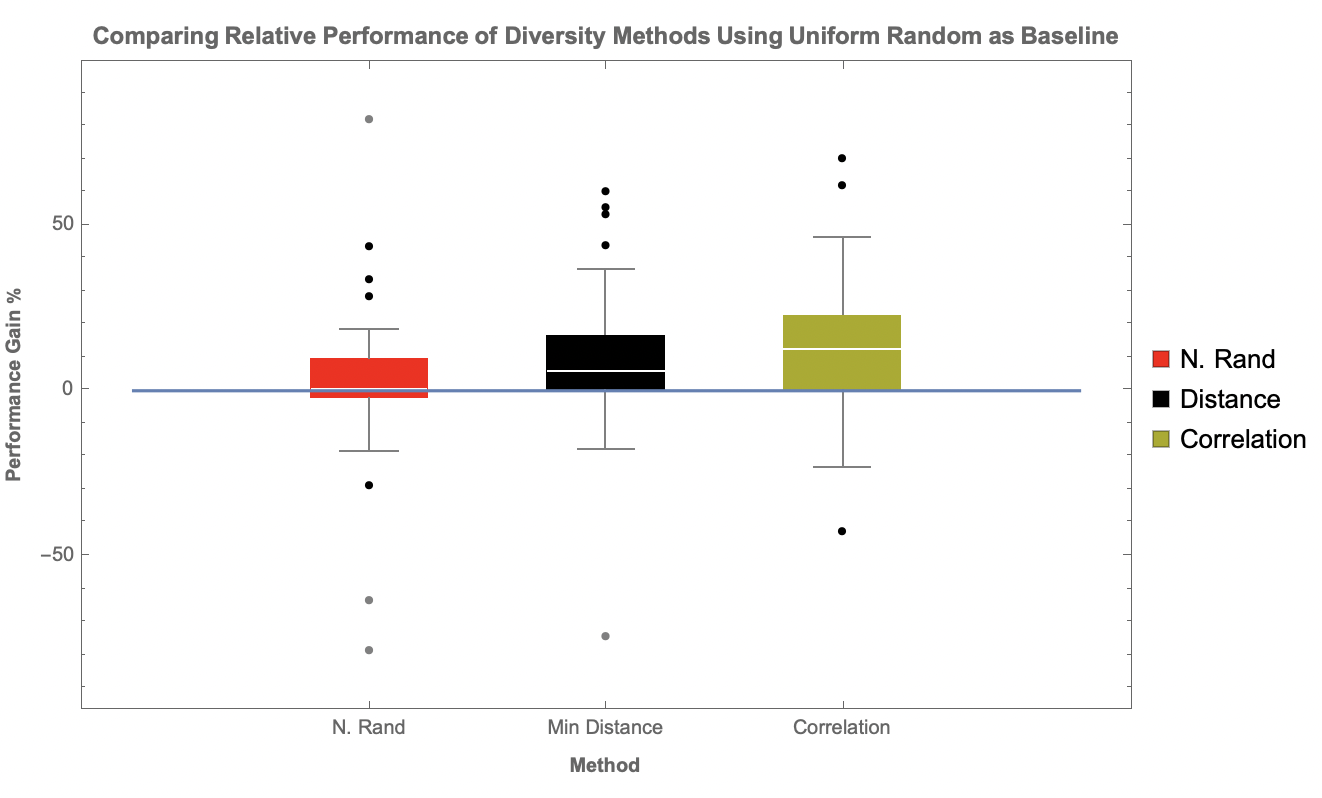}
\caption{{\bf Comparing Relative Performance of Diversity Methods Using Uniform Random Selection as Baseline.} Shown here are the performance differences of both the AL diversity methods compared to uniform random selection as the baseline (blue line) and normally distributed random selection (red distribution). We see that using minimum distance (green distribution) performs consistently better than the baseline and correlation (blue distribution) works best as a diversity metric. The drawback with using correlation as the diversity metric though is that it requires problems with more than two dimensions, so the problems with two dimensions are ignored when using correlation. The distributions represent the median performances of 100 independent runs across all test problems.}
\label{fig:dist}
\end{figure}

The results of comparing the different data diversity-based active learning methods are summarized in Figures \ref{fig:dist} and \ref{fig:distCount} and the full results are shown in Table \ref{tab:app1} in the Appendix.  Figure \ref{fig:dist} uses uniform random sampling as the baseline for comparison, shown as the blue line. We again include normally distributed random sampling for comparison as the red distribution. We can see that both diversity metrics have better performance than uniform random sampling, on average requiring fewer training points to find a solution. We also see that correlation as a diversity metric performs best, often requiring the least number of training data points to find a solution. Correlation does have the disadvantage, though, of not working on the problems with just two dimensions. Those two problems are not represented in the correlation bar in the chart since they are not applicable.

Figure \ref{fig:distCount} shows the number of cases where each method either outperformed or underperformed when compared to uniform random sampling. We see again that correlation has the best performance. This indicates that not only does correlation lead to requiring fewer training points on average, but also indicates that it most consistently requires fewer points. We see that distance as a metric requires fewer points than uniform and random sampling, but is not as consistent as correlation.

\begin{figure}[h]
\centering
\includegraphics[width=12cm]{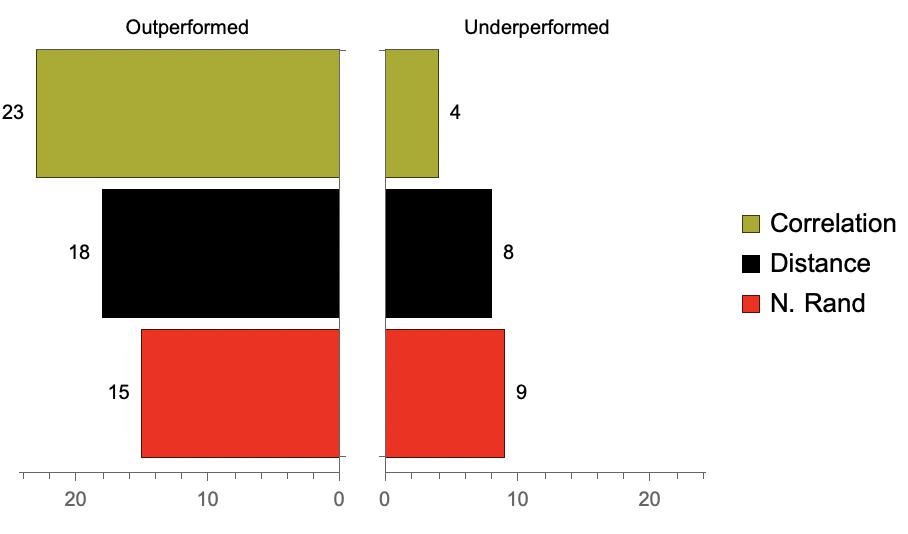}
\caption{{\bf Comparing Performance of Diversity Methods Against Uniform Random Selection.} Each method is compared to uniform random sampling and the number of times that the method outperforms and underperforms is reported. The number of times each method outperforms is shown on the left and the number of times each method underperforms is shown on the right. Outperforms means that a method used fewer points than uniform random sampling. Underperforms means it required more points. Ties are not counted but can be easily determined by taking the difference of 35 and the two values reported. The results show that correlation performed best, underperforming the fewest times and outperforming the most. }
\label{fig:distCount}
\end{figure}

\subsection{Comparing Diversity, Uncertainty, and Pareto Optimization of Both}
Next we explore how the performance compares when using uncertainty and diversity together to see if there are benefits to considering both for selecting training data with AL compared to just uncertainty or diversity alone.
For this comparison, we selected one diversity metric and one uncertainty metric. For the uncertainty metric, we chose differential entropy since it was shown to be the best performing metric in Figure \ref{fig:uncertainty}. For the diversity metric, we chose minimum distance. Although it didn't perform best, it is most versatile since it isn't restricted to problems with more than 2 dimensions. For the combination method, we used a Pareto optimization to find the points with the best trade-off of both the uncertainty and diversity metrics from 10,000 randomly generated points each iteration. From the Pareto front of points that are non-dominated in those two objectives, we ordered them based on their uncertainty score and selected the median point. Note that sorting based on uncertainty is just the reverse order of a sort by diversity, so which objective you choose to sort by shouldn't have a significant impact. The only impact would be on cases where an even number of points are on the front so the point you select isn't the true median but rather one of the points near the median. When this occurs, we round down to select the median point, which would give a slight bias toward uncertainty. By selecting the median point we are attempting to choose a point that has a relatively good balance between the two objectives. 

The results of this comparison are shown in Figures \ref{fig:tradeoff} and \ref{fig:tradeoffCount} with the results from each problem shown in the Appendix in Table \ref{tab:uncDiv}. Again in Figure \ref{fig:tradeoff}, we use uniform random sampling as the baseline (blue line) and include normally distributed random sampling for comparison. The results show that all three methods work better than the baseline and normally distributed random sampling. Using the uncertainty metric, differential entropy, works slightly better than using the distance metric, minimum distance. We also see that there is a benefit to combining both metrics using the Pareto optimization since we see an improvement in the upper quartile of performance. It is also interesting to note, as can be seen in Figure \ref{fig:tradeoffCount}, that the diversity metric alone performed worse than uniform random sampling in 8 of the 35 cases, whereas the uncertainty approach and the Pareto approach only performed worse in 4 of the cases, demonstrating that the uncertainty and Pareto approaches offer more consistent improvements. This indicates that it is important to consider the current models to help guide the AL process. This makes sense since the goal is to select training points that will best inform the current model population, using only diversity doesn't consider the current state of models, so it is less likely that the training points selected will most inform those models. Statistical significance tests were also performed and the number of cases determined to be statistically significant are shown in the darker regions in the figure. The Mann-Whitney test was used to test for significance and a threshold of 0.05 was used. The Pareto approach was found to be statistically significant in 18 of the 20 cases where the Pareto approach outperformed.

\begin{figure}[h]
\centering
\includegraphics[width=12cm]{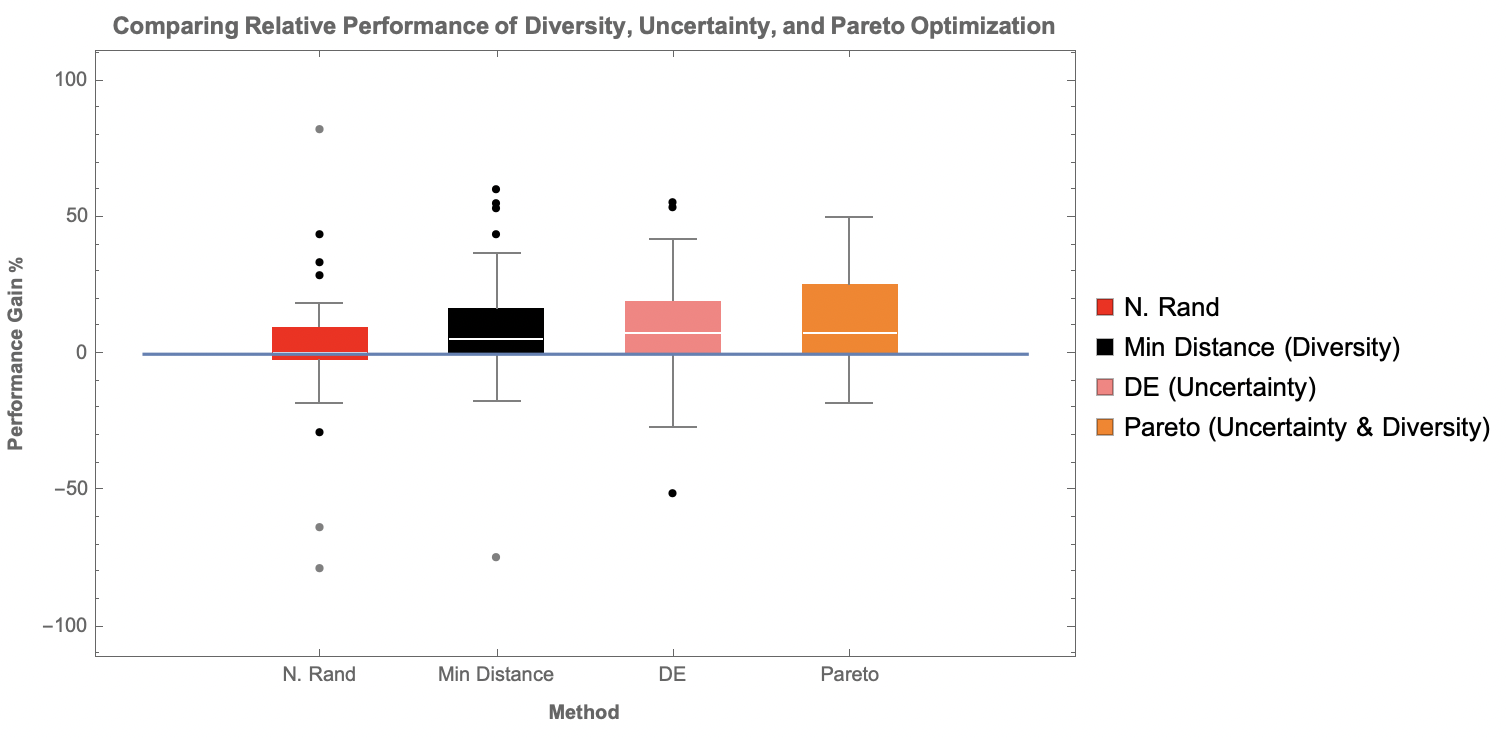}
\caption{{\bf Comparing Relative Performance of Diversity, Uncertainty, and Pareto Optimization Using Uniform Random Selection as Baseline.} Shown here are the performance differences of AL diversity, uncertainty and Pareto methods compared to uniform random selection as the baseline (blue line) and normally distributed random selection (red distribution). We see that using the diversity metric, minimum distance (green distribution), performs consistently better than the baseline and the uncertainty metric, DE (blue distribution), performs a bit better than the diversity method. When using a Pareto optimization of both diversity and uncertainty we get even better performance. The distributions represent the median performances of 100 independent runs across all test problems. For completeness, there is a single point around -150 for the Pareto approach.  }
\label{fig:tradeoff}
\end{figure}

\begin{figure}[h]
\centering
\includegraphics[width=12cm]{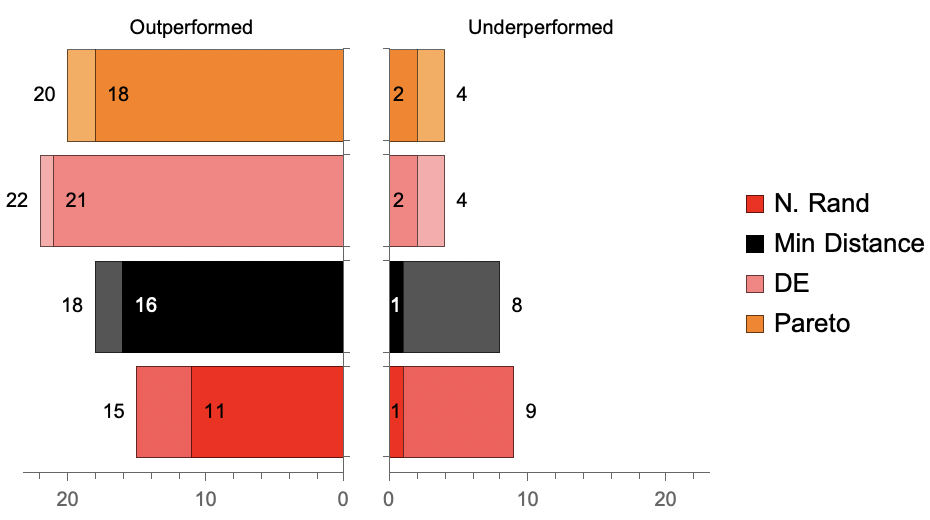}
\caption{{\bf Comparing Performance of Diversity, Uncertainty, and Pareto Optimization Against Uniform Random Selection.} Each method is compared to uniform random sampling and the number of times that the method outperforms and underperforms is reported. The number of cases where the differences are statistically significant is shown in the darker regions. The number of times each method outperforms is shown on the left and the number of times each method underperforms is shown on the right. Outperforms means that a method used fewer points than uniform random sampling. Underperforms means it required more points. Ties are not counted but can be easily determined by taking the difference of 35 and the two values reported. The results show that DE, the uncertainty method works best. The Pareto approach ties for the least number of underperforming cases, matching DE, and outperforms between DE and Min. Distance. Statistical significance was determined using the threshold of 0.05 with the Mann-Whitney test. }
\label{fig:tradeoffCount}
\end{figure}

 Looking at the results, there are two instances where the Pareto approach performed considerably worse than the uncertainty and diversity approaches. Those are equations 9 and 71. Table \ref{tab:uncDiv} in the Appendix shows that the combined method performs worse than focusing alone on either diversity or uncertainty for those two problems. This is likely a result of equations 9 and 71 being higher dimensional problems with 6 and 5 dimensions, respectively, so the 10,000 randomly generated points don't sufficiently fill the search space to find points with high values for both uncertainty and diversity. 

Equation 71 was further explored to see if sampling additional points improved the performance when using the combined diversity uncertainty approach and to verify that sparse sampling was at least part of the issue as suspected. Equation 71 was retested using 100,000 randomly sampled points to search for the best trade-off between diversity and uncertainty. When using 100,000 points the median number of points required to solve the problem decreased to 42 points from 50.5, confirming that better sampling of the space improves the performance in this higher dimensional problem. The median performance of 42 points is still worse than either of the uncertainty or diversity approaches, so more points could be used, but increasing the number of points beyond 100,000 begins to make that search rather expensive. Rather than randomly sampling the points then selecting the Pareto front from those points, an alternative optimization method, such as NSGA II \citep{nsga}, could be used in future studies which might be cheaper and likely more effective. 

\subsection{Additional Benchmark Problems}

To further test the Pareto AL approach, we selected two problems from a more recent benchmark set, SRBench \citep{SRBench}. 
One that is on the easier side for StackGP and one that is a bit more challenging. The easier problem selected was the van der Pol oscillator problem, referred to as "strogatz\_vdp1" in SRBench. The equation for the van der Pol oscillator problem that we are trying to rediscover is $x' = 10 *  (y - (1)/(3) * (x^3-x))$. The more challenging problem was the bar magnet problem, referred to as "strogatz\_barmag1" in SRBench and the equation for the bar magnet problem that we are trying to rediscover is $x'=0.5*sin(x-y)-sin(x)$.  
As with the previous problems, we performed each experiment 100 times and computed the median number of points to find the solution. The results of those experiments are shown in Table \ref{tab:SRBench}. We can see that the Pareto approach performs significantly better than randomly sampling from a normal distribution and performs about 27.8\% better than randomly sampling from a uniform distribution on the bar magnet problem. The performance gains over the normal and uniform distributed samplings are statistically significant considering a threshold of 0.05 using the Mann-Whitney test. We computed a p-value of \text{3.490$*{10}^{-11}$} when comparing to the normal distribution and \text{6.481$*{10}^{-6}$} when comparing to the uniform sampling. We also see better performance on the van der Pol oscillator, although since it was an easy problem there isn't as much opportunity for improvement, so we only see a reduction of a few points. The performance gains over the normal and uniform distributions are again statistically significant with a p-value of \text{2.51$*{10}^{-7}$} when compared with the results from using normally distributed sampling and a p-value of \text{4.008$*{10}^{-13}$} when compared with the results from using uniform random sampling.

\begin{table}[h!]
\centering
\small

\label{tab:SRBench}
\begin{tabular}{cccl} 
 \toprule
 \rowcolor{gray!50}
 SRBench  & N. Ran & U. Ran & Pareto AL \\ [0.5ex]
 \rowcolor{gray!50}
 Problem & Data Pts. & Data Pts.  & Data Pts.\\ [0.5ex]
 \midrule
 Bar Magnet \#1 & 51 & 18 & 13 \\
 \hline
 Van der Pol Osc. \#1 & 10 & 9 & 7 \\
  \bottomrule
 \end{tabular}
 \caption{Shown are the median numbers of points needed to solve each equation. A total of 100 independent trials were performed for each equation. We compare the active learning method that uses both diversity and uncertainty and compare the performance against random sampling on two problems from the SRBench. }
\end{table}

\section{Conclusion}

Both uncertainty and diversity metrics for active learning were explored to see how each metric impacts the success of active learning in genetic programming. As well, a Pareto approach was defined that allows both diversity and uncertainty to be considered for active learning. Of the uncertainty approaches, it was observed that differential entropy performed best. It was also observed that relative uncertainty functions did not perform well. When using differential entropy it was found that performance could be boosted by using differential evolution as the optimizer over Scipy Optimize's minimize function. This indicates that the search space is not convex and requires a good optimizer to find solutions with high uncertainty. 

When comparing the data diversity methods, it was found that correlation performed better than minimum Euclidean distance. Although correlation worked better, it does not work on cases with 2 dimensions or less. Thus, minimum Euclidean distance was selected for the Pareto approach. Future implementations may default to using minimum Euclidean distance for all cases with 1 or 2 dimensions and using correlation for higher dimensional problems. Mean distance was considered, but determined to be uninformative due to its frequency of identifying repeat points. 

When comparing the Pareto approach which used both differential entropy and minimum Euclidean distance to differential entropy, minimum Euclidean distance, uniform random selection, and normally distributed random selection, it was found that differential entropy worked best, with the Pareto approach performing between differential entropy and minimum Euclidean distance. Looking at individual problems, there were a few cases where the Pareto approach actually worked better than both differential entropy and minimum Euclidean distance on their own, indicating potential benefits of combining the two approaches. For the cases where the Pareto approach did not work as well, it was identified that the multi-objective optimization strategy may have been at fault since it relies on randomly generating N points and selecting the median value in the Pareto front. Better methods such as NSGA-II could be explored in future studies to see if improved optimization methods leads to better active learning performance. 

Overall, it was found that active learning can be efficiently utilized with genetic programming to reduce training data requirements. In practice, this would be useful to apply in scenarios where collecting data or labelling data is expensive, and model training is relatively cheap. In these scenarios, active learning could be used to guide data collection and labelling so that good models can be arrived at using as few data points as possible. This application has the potential to accelerate data driven research, since it could lead to finding solutions with fewer resources in less time.

\bmhead{Acknowledgments}

Computer support by MSU's iCER high-performance computing center is gratefully acknowledged.

\bmhead{Data availability} The datasets generated/analysed during the current study are available from the corresponding author on reasonable request.
\bmhead{Code availability} The code for StackGP with active learning can be found here: https://github.com/hoolagans/StackGP

\section{Appendix}


\begin{table*}[h]
\centering
\tiny

\begin{tabular}{|c|c|c|c|c|c|c|c|l|} 
 \hline
 \rowcolor{gray!50}
 EQ  & U. Ran & N. Ran & Std/Mean & TrStd/TrMean & Std/TrMean & Std & DE & DE (DE) \\ [0.5ex]
 \rowcolor{gray!50}
 Num & Data & Data & Data & Data  & Data & Data & Data & Data\\ [0.5ex]
 \hline
 2 & 54.5 & 97.5  & 50 & 39 & 47 & 53 & 82.5 & 47 \\
 \hline
 3 & $>1000$& $>1000$ & 876  & 692  & 741 & $>1000$  &  $>1000$ & 724.5 \\
 \hline
 4 & 30 & 21.5  & 21.5 & 20 & 23 & 20 & 28 & {\bf 19.5}\\
 \hline
 7 & 88.5 & 82  & 23 & {\bf 21} & 22.5 & 39 & 52.5 & 35 \\
 \hline
 9 & 120.5 & 155.5  & 150.5 & 73.5 & 359.5 & 100.5 & 153 & 160 \\
 \hline
 10 & 6 & 6  & 6 & 11 & 6 & 7 & 6 & 6\\
 \hline
 13 & 13 & 14.5  & 15 & 15 & 14 & 14 & {\bf 12} & {\bf 12} \\
 \hline
 14 & 30.5 & 33 & 28 & 24 & 31 & 23.5 & 24 & 22.5\\
 \hline
 23 & 8 & 8  & 7 & 8 & 7 & 8 & 7.5 & 7\\
 \hline
 24 & 49 & 58  & 39 & 29.5 & 31 & 26 & {\bf 22} & 28\\
 \hline
 27 & 30 & 17 & 20 & 13 & 19.5 & 18 & 14 & 15\\
 \hline
 32 & 17 & 16 & 20 & 18 & 21 & 16 & 18 & {\bf 12}\\
 \hline
 35 & 19 & 17 & 17 & {\bf 6} & 21 & 18 & 13.5 & 12.5\\
 \hline
 39 & 10 & 10 & 10 & 12 & 11 & 10 & 9 & 9 \\
 \hline
 41 & 7 & 7 & 7 & 8 & 7 & 8 & 7 & 7 \\
 \hline
 43 & 453 & 82  &  876 & 218 & 202.5  &  144 &  326 & 192.5 \\
 \hline
 47 & 13 & 12 & 14 & 12 & 13 & 13 & 12 & 12 \\
  \hline
 48 & 15.5 & 14 & 18 & 17 & 17.5 & 14 & 13 & {\bf 12.5} \\
  \hline
 52 &  9.5 & 9 & 10 & 10 & 9.5 & 10 & 9 & 9 \\
  \hline
 55 &  10 & 10 & 11 & 12 & 10 & 11 & 10 & {\bf 9} \\
  \hline
 57 & 30.5 & 31.5  & 25.5 & 27 & 24 & {\bf 17} & 23  &  21 \\
  \hline
 60 &  7 & 7 & 7 & 7 & 7 & 7 & 7 & 7 \\
  \hline
 61 & 18.5 & 20  & 20 & 19 & 18 & 18 & {\bf 16} & 17 \\
  \hline
 62 & 34.5 & 56.5 & 37.5 & 34.5 & 33 & 34 & 30 & {\bf 28.5} \\
  \hline
 63 & 14 & 13  & 15 & 16 & 15.5 & 14 & 13 & 13 \\
  \hline
 66 & 11 & 9  & 15 & 14 & 14 & 15 & 10 & {\bf 9} \\
  \hline
 67 & 10.5 & 11  & 11 & 10 & 10 & 10 & 11 & 11\\
  \hline
 71 & 51.5 & 47 & 34 & 58 & 38.5 & 31 & \bf{30} & 35 \\
  \hline
 83 & 5 & 5 & 5 & 5 & 5 & 5 & 5 & 5 \\
  \hline
 85 & 4 & 4 & 4 & 4 & 4 & 4 & 4 & 4 \\
  \hline
 89 & 5 & 5 & 4 & 5 & 4 & 4.5 & 5 & 5 \\
  \hline
 93 & 8 & 8 & 8 & 8 & 8 & 7 & 8 & 8 \\
  \hline
 95 & 11 & {\bf 9} & 12 & 11.5 & 11 & 12 & 10 & 11 \\
  \hline
 98 & 8 & 7 & 9 & 9 & 9 & 9 & 7 & 7.5 \\
  \hline
 99 & 30 & {\bf 20}  & 31 & 35.5 & 35 & 25 & 24 & 21 \\

  \hline
\rowcolor{gray!0}
  Vs. & & & & & & & & \\ \rowcolor{gray!0} U. Sampl. & - & - & {\bf 19} & {\bf 20} & {\bf 24}  & {\bf 27} & {\bf 31} & {\bf 33} \\
 \hline
 \rowcolor{gray!25}
 Vs. & & & & & & & & \\N. Sampl. & - & - & {\bf 22} & {\bf 20} & {\bf 20} & {\bf 22} & {\bf 29} & {\bf 30}\\
 \hline
\end{tabular}
\caption{Shown are the median number of points needed to solve each equation. A total of 100 independent trials were performed for each equation. The last row indicates the number of cases where each of the active learning methods matched or performed better than uniform random sampling of training data. Where informative, the minimum number is bolded. }
\label{tab:uncTab}
\end{table*}

\begin{table}[h]
\centering
\small

\begin{tabular}{cccl} 
 \toprule
 \rowcolor{gray!50}
 EQ  & U. Rand & Pt. Dist & Pt. Corr \\ [0.5ex]
 \rowcolor{gray!50}
 Num & Data Pts. & Data Pts.  & Data Pts.\\ [0.5ex]
 \midrule
 2 & 54.5 & \textbf{44} & -\\
 \hline
 3 & $>1000$ & $>1000$ & $>1000$ \\
 \hline
 4 & 30 & \textbf{19} & 29.5\\
 \hline
 7 & 88.5 & 35.5 & 60\\
 \hline
 9 & 120.5 & 210.5 & \textbf{102.5}\\
 \hline
 10 & 6 & 6 & \textbf{5}\\
 \hline
 13 & 13 & \textbf{12} & 14\\
 \hline
 14 & 30.5 & 24 & \textbf{22}\\
 \hline
 23 & 8 & \textbf{7} & 8\\
 \hline
 24 & 49 & \textbf{23} & 26.5\\
 \hline
 27 & 30 & 13.5 & \textbf{11.5}\\
 \hline
 32 & 17 & 15.5 & \textbf{14}\\
 \hline
 35 & 19 & \textbf{13.5} & 15\\
 \hline
 39 & 10 & 11 & \textbf{9}\\
 \hline
 41 & 7 & 8 & \textbf{6}\\
 \hline
 43 & 453 & 533.5 & \textbf{136.5}\\
 \hline
 47 & \textbf{13} & 14.5 & 16\\
  \hline
 48 & 15.5 & \textbf{13} & \textbf{13}\\
  \hline
 52 & 9.5 & 10 & \textbf{9}\\
  \hline
 55 & 10 & 10 & 10\\
  \hline
 57 & 30.5 & \textbf{28} & 29.5\\
  \hline
 60 & 7 & 7 & 7\\
  \hline
 61 & 18.5 & \textbf{17.5} & \textbf{17.5}\\
  \hline
 62 & 34.5 & \textbf{29.5} & 36.5\\
  \hline
 63 & 14 & 14 & \textbf{12}\\
  \hline
 66 & 11 & 12 & \textbf{10}\\
  \hline
 67 & \textbf{10.5} & 11 & 15\\
  \hline
 71 & 51.5 & \textbf{29} & 30.5\\
  \hline
 83 & 5 & 5 & 5\\
  \hline
 85 & 4 & 4 & -\\
  \hline
 89 & 5 & 5 & 5\\
  \hline
 93 & 8  & 7.5 & \textbf{7}\\
  \hline
 95 & 11 & 10 & \textbf{8}\\
  \hline
 98 & 8 & 8 & \textbf{7}\\
  \hline
 99 & 30 & 25 & \textbf{20.5}\\

  \bottomrule
 Perf. Count  &  -  & 27/35 & 29/33 \\
\end{tabular}
\caption{Shown are the median number of points needed to solve each equation. A total of 100 independent trials were performed for each equation. There are 2 equations that have a dash instead of a number and that is because they have only two dimensions, so selecting points with minimal correlation to the rest of the training set is not possible. The approach using uniformly random data points was included in the first column represented as a baseline. The last row indicates the number of cases where each of the point diversity methods matched or performed better than the random approach.}
\label{tab:app1}
\end{table}
\begin{table}[h]
\centering
\small

\begin{tabular}{cccl} 
 \toprule
 \rowcolor{gray!50}
 EQ  & Pt. Dist & Pareto & Pt. Unc. \\ [0.5ex]
 \rowcolor{gray!50}
 Num & Data Pts. & Data Pts.  & Data Pts.\\ [0.5ex]
 \midrule
 2 & 44 & 36.5 & 82.5 \\
 \hline
 3 & $>1000$ & 501 & $>1000$ \\
 \hline
 4 & 19 & 29 & 28\\
 \hline
 7 & 35.5 & 48.5 & 52.5\\
 \hline
 9 & 210.5 & 304 & 153\\
 \hline
 10 & 6 & 6 & 6\\
 \hline
 13 & 12 & 12 & 12\\
 \hline
 14 & 24 & 22 & 24\\
 \hline
 23 & 7 & 8 & 7.5\\
 \hline
 24 & 23 & 27 & 22\\
 \hline
 27 & 13.5 & 19 & 14\\
 \hline
 32 & 15.5 & 14 & 18\\
 \hline
 35 & 13.5 & 14 & 13.5\\
 \hline
 39 & 11 & 10 & 9\\
 \hline
 41 & 8 & 7 & 7\\
 \hline
 43 & 533.5 & 314.5 & 326\\
 \hline
 47 & 14.5 & 12 & 12\\
  \hline
 48 & 13 & 13 & 13\\
  \hline
 52 & 10 & 10 & 9\\
  \hline
 55 & 10 & 9 & 10\\
  \hline
 57 & 28 & 24 & 23\\
  \hline
 60 & 7 & 7 & 7\\
  \hline
 61 & 17.5 & 15 & 16\\
  \hline
 62 & 29.5 & 21.5 & 30\\
  \hline
 63 & 14 & 14 & 13\\
  \hline
 66 & 12 & 10 & 10\\
  \hline
 67 & 11 & 11 & 11\\
  \hline
 71 & 29 & 50.5 & 30\\
  \hline
 83 & 5 & 5 & 5\\
  \hline
 85 & 4 & 4 & 4\\
  \hline
 89 & 5 & 5 & 5\\
  \hline
 93 & 7.5  & 8 & 8\\
  \hline
 95 & 10 & 13 & 10\\
  \hline
 98 & 8 & 8 & 7\\
  \hline
 99 & 25 & 25.5 & 24\\

  \bottomrule
 Worst Count  & 13 & 11 & 8 \\ 
 Best Count  & 16 & 19 & 21 \\
\end{tabular}
\caption{Shown are the median number of points needed to solve each equation. A total of 100 independent trials were performed for each equation. Here the trade-off between diversity and uncertainty is explored. The second to last row indicates the number of times each approach was the worst of the three approaches. The last row indicates the number of cases where each approach was the best or tied for the best of the three approaches. Minimum point distance was used for the diversity metric and differential entropy was used as the uncertainty metric.}
\label{tab:uncDiv}
\end{table}

\begin{table}[!t]
\centering
\small

\begin{tabular}{cccl} 
 \toprule
 \rowcolor{gray!50}
 EQ  & p-value & Significant \\ [0.5ex]
 \rowcolor{gray!50}
 Num &  & \\ [0.5ex]
 \midrule
 2 &9.18077$*{10}^{-4}$& Yes \\
 \hline
 3 & \text{1.4012505382465074$*{10}^{-11}$} & Yes \\
 \hline
 4 &0.612865 & No \\
 \hline
 7 & \text{9.606121209101481$*{10}^{-6}$} & Yes \\
 \hline
 9 &1.78579$*{10}^{-2}$ & Yes \\
 \hline
 10 &3.40099$*{10}^{-5}$& Yes \\
 \hline
 13 &2.38254$*{10}^{-5}$& Yes \\
 \hline
 14 &7.46075$*{10}^{-4}$ & Yes \\
 \hline
 23 & \text{3.3175302779045554$*{10}^{-6}$} & Yes \\
 \hline
 24 &8.64911$*{10}^{-5}$ & Yes \\
 \hline
 27 &0.0354696 & Yes \\
 \hline
 32 &1.30021$*{10}^{-3}$ & Yes \\
 \hline
 35 &2.90356$*{10}^{-3}$  & Yes \\
 \hline
 39 &7.61664$*{10}^{-3}$  & Yes \\
 \hline
 41 & \text{5.872949580996268$*{10}^{-7}$} & Yes \\
 \hline
 43 &3.38224$*{10}^{-3}$  & Yes \\
 \hline
 47 &1.91012$*{10}^{-4}$ & Yes \\
  \hline
 48 &1.78267$*{10}^{-5}$ & Yes \\
  \hline
 52 & 0.433203 & No \\
  \hline
 55 &1.98428$*{10}^{-4}$ & Yes \\
  \hline
 57 &2.51679$*{10}^{-5}$ & Yes \\
  \hline
 60 &6.84394$*{10}^{-3}$ & Yes \\
  \hline
 61 &1.39827$*{10}^{-3}$ & Yes \\
  \hline
 62 &\text{2.6897830944058467$*{10}^{-20}$} & Yes \\
  \hline
 63 & 0.0399119 & Yes \\
  \hline
 66 &1.55871$*{10}^{-3}$ & Yes \\
  \hline
 67 & 0.245499 & No \\
  \hline
 71 & 0.392183 & No \\
  \hline
 83 & \text{9.35933923552931$*{10}^{-10}$} & Yes \\
  \hline
 85 & \text{1.059722812150408$*{10}^{-12}$} & Yes \\
  \hline
 89 & 0.140942 & No \\
  \hline
 93 &1.54882$*{10}^{-5}$ & Yes \\
  \hline
 95 & 0.0345741 & Yes \\
  \hline
 98 &2.98341$*{10}^{-3}$ & Yes \\
  \hline
 99 &5.63906$*{10}^{-3}$ & Yes \\

  \bottomrule
 Significance Count  &  \textbf{30/35} &\\
\end{tabular}
\caption{Statistical significance of Pareto AL approach vs. uniform random sampling. We are using a threshold of 0.05 to test for significance. The Mann-Whitney test was used to test for significance. }
\label{tab:sig1}
\end{table}





\clearpage
\bibliography{sn-article}

\end{document}